\documentclass{article}

\usepackage[final]{neurips_2021}

\usepackage[utf8]{inputenc}
\usepackage[T1]{fontenc}
\usepackage{hyperref}
\usepackage{url}
\usepackage{booktabs}
\usepackage{amsfonts}
\usepackage{nicefrac}
\usepackage{microtype}
\usepackage{xcolor}
\usepackage{paralist}
\usepackage{graphicx}
\usepackage{tabularx}

\newcolumntype{L}[1]{>{\raggedright\arraybackslash}p{#1}}
\newcolumntype{C}[1]{>{\centering\arraybackslash}p{#1}}
\newcolumntype{R}[1]{>{\raggedleft\arraybackslash}p{#1}}

\title{Learning to Transfer for Traffic Forecasting via Multi-task Learning}

\author{
  Yichao Lu \\
  Layer 6 AI \\
  \texttt{yichao@layer6.ai} \\
}

\begin{document}

\maketitle

\begin{abstract}
Deep neural networks have demonstrated superior performance in short-term traffic forecasting. However, most existing traffic forecasting systems assume that the training and testing data are drawn from the same underlying distribution, which limits their practical applicability. The NeurIPS 2021 Traffic4cast challenge is the first of its kind dedicated to benchmarking the robustness of traffic forecasting models towards domain shifts in space and time. This technical report describes our solution to this challenge. In particular, we present a multi-task learning framework for temporal and spatio-temporal domain adaptation of traffic forecasting models. Experimental results demonstrate that our multi-task learning approach achieves strong empirical performance, outperforming a number of baseline domain adaptation methods, while remaining highly efficient. The source code for this technical report is available at \url{https://github.com/YichaoLu/Traffic4cast2021}.
\end{abstract}

\section{Introduction}

Traffic forecasting is a crucial task in intelligent transportation systems aimed at estimating the future traffic flows based on historical observations [1]. Advanced traffic forecasting systems are of great social and economic significance [2]. However, traffic forecasting is particularly challenging due to the complex spatio-temporal dependencies among traffic flows [3]. Compared to other spatio-temporal learning tasks such as video semantic segmentation [4] and precipitation nowcasting [5], the research on traffic forecasting remains quite limited, primarily due to the lack of a large-scale, authoritative benchmark dataset. 

The Traffic4cast competition series \footnote{https://www.iarai.ac.at/traffic4cast/} hosted by the \textit{Institute of Advanced Research in Artificial Intelligence (IARAI)} \footnote{https://www.iarai.ac.at/} is dedicated to benchmarking existing traffic forecasting systems and to advancing the state-of-the-art. The dataset used in the challenge is derived from the GPS trajectories of a large fleet of probe vehicles, made available by \textit{HERE Technologies} \footnote{https://www.here.com/}.

Deep neural networks have demonstrated exceptional performance in the Traffic4cast challenges at NeurIPS 2019 [6] and NeurIPS 2020 [7]. However, the training and testing data for the Traffic4cast 2019 and Traffic4cast 2020 challenges are drawn from the same underlying distribution, which limits the practical applicability of the winning solutions.

Going beyond the success at NeurIPS 2019 and NeurIPS 2020, Traffic4cast 2021 poses additional challenges for participants. In particular, it seeks to capture the underlying patterns of traffic flow that are both robust and transferable. The robustness of the solutions is benchmarked through \begin{inparaenum}[(i)]
\item the core challenge, where models need to adapt to a drastic temporal domain shift due to the Covid-19 pandemic, and 
\item the extended challenge, where models need to predict the traffic flow in entirely new cities.
\end{inparaenum}
Participants are therefore asked to build robust traffic forecasting systems that can adapt to domain shifts in space and time. 

In this technical report, we present a multi-task learning approach to this challenge. We demonstrate that a baseline U-Net model, when trained in a multi-task learning setup, achieves strong empirical performance in both temporal and spatio-temporal domain adaptation. Extensive experiments in the NeurIPS 2021 Traffic4cast challenge show that the multi-task learning approach outperforms a number of state-of-the-art domain adaptation methods. We additionally discuss a few possible avenues of further improvements from a practitioner's perspective.

\begin{figure}[!t]
    \centering
    \includegraphics[width=\textwidth]{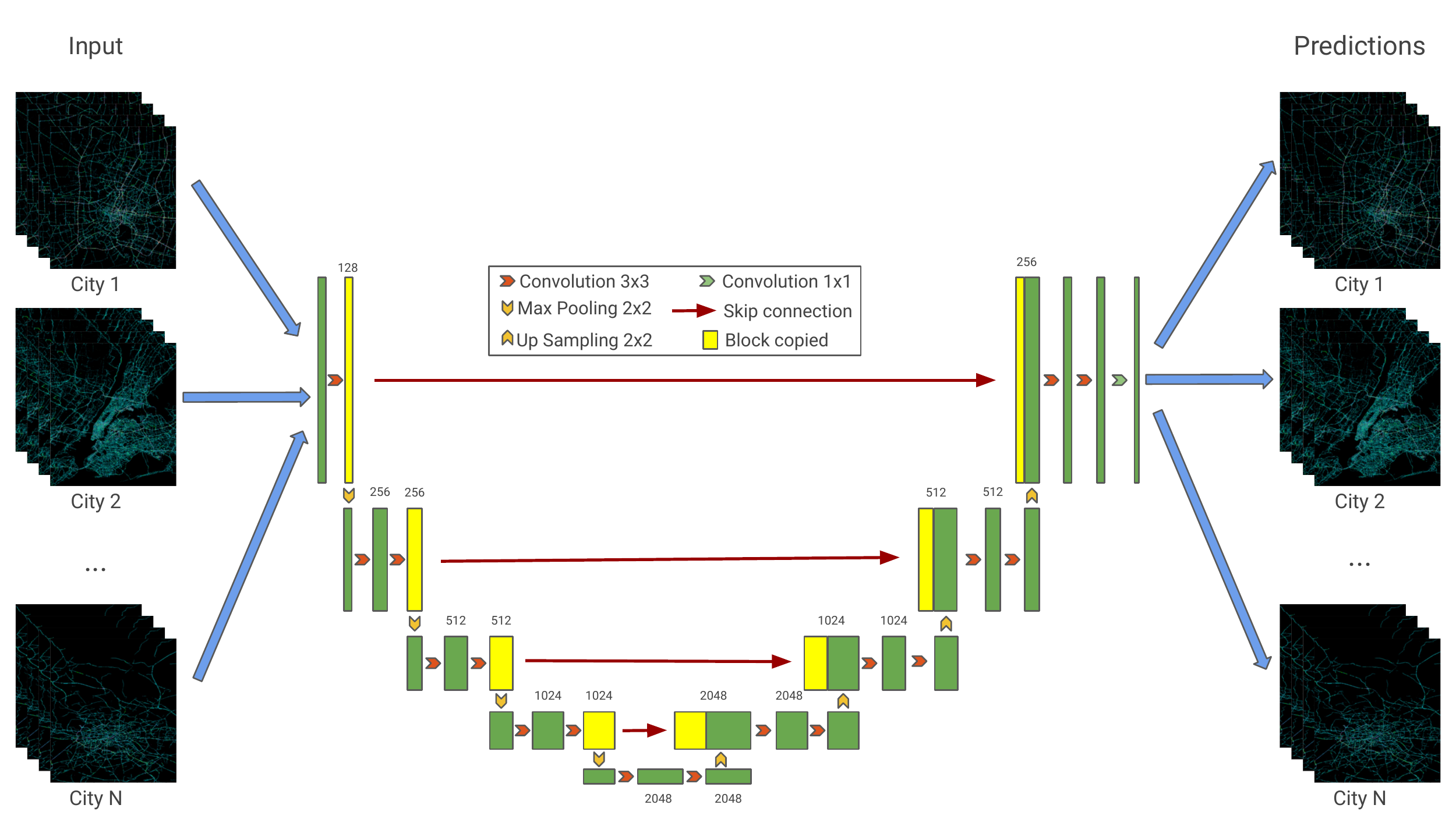}
    \caption{A multi-task learning framework for traffic forecasting.}
    \label{fig:arch}
\end{figure}

\section{Related work}

\subsection{Traffic forecasting}

Recent years have witnessed a surge of interest in traffic forecasting, primarily due to its potential of application in intelligent transportation systems [8]. Conventional traffic forecasting models rely heavily on classic statistical learning methods, e.g., autoregressive integrated moving average (ARIMA) [9], wavelet transform [10], and radial basis function network [11]. 

Lately deep neural networks have enjoyed tremendous success in a variety of application domains due to their ability to capture complex dependencies and non-linearities in large datasets. Unsurprisingly, the traffic forecasting community has recently turned its attention towards utilizing deep neural networks [12, 13, 14]. A proven effective approach is to represent the dynamics of a complex traffic system as a sequence of movie frames, where each pixel corresponds to the traffic intensity at a certain block of area and each frame summarizes a discrete time bin, thus casting traffic forecasting as a video prediction task [6, 7].

Many of the early works focused on convolutional neural networks which are mainly used for visual tasks such as image classification and object detection. In order to model temporal relationships using a convolutional architecture, a common practice is to concatenate the sequence of frames along the channel dimension and to apply 2-dimensional convolutional neural networks [15]. Another approach is 3-dimensional convolutional neural networks, where convolution and pooling operations are performed spatio-temporally [16].

In another stream of works, researchers exploited the potential of recurrent neural networks to capture temporal dependencies. While recurrent neural networks are by design capable of capturing non-linear time-dependent dynamics, vanilla recurrent units suffer from gradient vanishing and exploding problems and popular variants such as the Long Short-Term Memory (LSTM) [17] and the Gated Recurrent Unit (GRU) [18] are often used [19].
Many state-of-the-art traffic forecasting models have also employed a hybrid of convolutional and recurrent layers as the underlying architecture [14, 20, 21, 22]. Such architectural design enables the model to simultaneously exploit the ability of convolutional units to model spatial relationships and the potential of recurrent units to capture temporal dependencies.

Convolutional neural networks and recurrent neural networks have the limitation that they work only on data with an underlying Euclidean or grid-like structure, and, as such, fail to capture the complex graph structures in transportation systems such as the road network [23]. Recently graph neural networks have shown exceptional performance in a variety of traffic flow prediction tasks due to their ability to capture spatial dependencies presented in the form of non-Euclidean graph structures [24, 25, 26]. It has also been demonstrated that graph neural networks can more effectively learn properties given by the underlying road network, which improves the generalization performance when making predictions on previously unseen cities [27].

\subsection{Domain adaptation}

While deep neural networks have achieved state-of-the-art results in a variety of machine learning tasks, they typically assume that the training and testing data are drawn from the same distribution. This assumption may not hold in real-world scenarios, and hence the need for domain adaptation [28, 29]. 

The intuition behind domain adaptation is to learn domain-invariant representations that also lead to strong performance in the source domain [30]. Deep domain adaptation approaches can be categorized into three groups [31]: \begin{inparaenum}[(i)]
\item discrepancy-based domain adaptation approaches that minimize the distance between the empirical source and target mapping distributions to reduce the domain shift [32, 33, 34],
\item adversarial-based domain adaptation approaches that employ domain discriminators to encourage domain confusion via an adversarial objective [35, 36], and 
\item reconstruction-based domain adaptation approaches that use data reconstruction as an auxiliary task to ensure feature invariance [37].
\end{inparaenum}
It has also been demonstrated that multi-task learning is an effective approach to domain adaptation [38, 39]. Multi-task learning can be viewed as a discrepancy-based domain adaptation approach in the sense that it implicitly minimizes the pairwise discrepancy between all the tasks [40].

\begin{table}[!t]
\centering
\begin{tabular}{L{0.38\textwidth}|C{0.2\textwidth}|C{0.22\textwidth}}
Method & MSE & Training time (hours) \\ \hline \hline
Naive Average & 53.406 & - \\ \hline
U-Net [41] & 49.127 & 24.8 \\ \hline
Graph ResNet [27] & 49.546 & 48.4 \\ \hline
U-Net + AdaBN [48] & 49.257 & 24.8 \\ \hline
U-Net + DaNN [32] & 49.096 & 26.4 \\ \hline
U-Net + DDC [33] & 49.223 & 28.8 \\ \hline
U-Net + DeepCORAL [34] & 49.230 & 26.8 \\ \hline
U-Net + ADDA [35] & 49.176 & 27.6 \\ \hline
U-Net + DANN [36] & 49.104 & 26.8 \\ \hline
U-Net + DSN [37] & 49.072 & 26.4 \\ \hline
U-Net + MAML [50] & 49.054 & 68.4 \\ \hline
\textbf{U-Net + Multi-task Learning (ours)} & \textbf{48.659} & \textbf{20.5} \\ \hline
\end{tabular}
\caption{Performance of our multi-task learning approach and the baseline methods in the core competition.}
\label{table:core_1}
\end{table}

\section{Approach}

In this section, we present our multi-task learning framework for traffic forecasting. We begin by introducing our baseline traffic forecasting model based on the U-Net architecture [41]. After that, we propose a multi-task learning approach to handling the temporal and spatio-temporal domain shifts in traffic forecasting. Finally, we describe the implementation details of our model.

\begin{table}[!t]
\centering
\begin{tabular}{C{0.2\textwidth}|C{0.2\textwidth}|C{0.2\textwidth}}
Rank & Team & MSE \\ \hline \hline
\textbf{1} & \textbf{oahciy (ours)} & \textbf{48.422} \\ \hline
2 & sungbin & 48.494 \\ \hline
3 & sevakon & 49.379 \\ \hline
4 & ai4ex & 49.720 \\ \hline
5 & Bo & 50.219 \\ \hline
\end{tabular}
\caption{Core competition leaderboard results for the top $5$ teams.}
\label{table:core_2}
\end{table}

\subsection{U-Net for traffic forecasting}

We employ U-Net as our baseline model due to its demonstrated superior performance in short-term traffic forecasting. Initially proposed for biomedical image segmentation, U-Net has been further adopted in a variety of pixel-level dense prediction tasks. The U-Net architecture consists of a contracting path (an encoder) and a symmetric expanding path (a decoder), thus yielding a U-shaped structure; see Figure \ref{fig:arch}. The encoder consists of a sequence of $K$ blocks, where each block is composed of two $3 \times 3$ convolutional layers, each followed by a group normalization layer [42] with group size set to $8$ and a rectified linear activation unit, and a $2 \times 2$ max-pooling layer with stride $2$. The number of filters in the convolutional layers is doubled after each downsampling operation. The decoder also consists of a sequence of $K$ blocks, where each block is composed of two $3 \times 3$ convolutional layers and a $2 \times 2$ upsampling layer, which performs the transposed convolution operation and halves the number of filters. We apply two $3 \times 3$ convolutional layers as the bridge to connect the encoder with the decoder, and apply a $1 \times 1$ convolutional layer as the final layer to generate the future traffic flow predictions.

The input to the traffic forecasting model is a $12 \times 495 \times 436 \times 8$ tensor, where $495 \times 436$ is the spatial resolution of the city heatmap, and the channel values of each pixel encode the observed traffic in each $100m \times 100m$ grid within a $5$ minute interval. The $8$ channels of each heatmap correspond to the volume and speed for each of the four headings (\textit{NE}, \textit{SE}, \textit{SW}, and \textit{NW}), where the values are normalized and discretized to an integer number between $0$ and $255$. The input to the traffic forecasting model consists of a stack of $12$ consecutive heatmaps of 5 minute interval time bins, spanning a total of $1$ hour. 

We concatenate the 12 heatmaps across the channel dimension, which results in a tensor of shape $495 \times 436 \times 96$. We additionally concatenate the static information of shape $495 \times 436 \times 9$ to the input, where the $9$ channels of the static information encode the density of the road network and the road connections to the $8$ neighboring cells. U-Net takes as input the $495 \times 436 \times 105$ tensor, and outputs a tensor of shape $495 \times 436 \times 48$. The output from U-Net is then reshaped into $6 \times 495 \times 436 \times 8$, where the $6$ predicted heatmaps correspond to the predicted traffic states $5$, $10$, $15$, $30$, $45$ and $60$ minutes in the future, respectively.

\subsection{A multi-task learning approach}

We present a multi-task learning framework for traffic forecasting; see Figure \ref{fig:arch}. Generally speaking, during training, instead of feeding the model with data from a particular city, we randomly sample data from all the available cities, and train the model to jointly predict the future traffic states for different cities. Assume that we have $M$ cities and the $i$-th city has $N_{i}$ training instances, the objective function of the multi-task learning framework is to simultaneously minimize the pixel-wise squared difference between the predicted traffic map movies and the ground truths across all cities:

\begin{equation}
    MSE = \frac{1}{M}\sum_{i = 1}^{M} \frac{1}{N_{i}} \sum_{j=1}^{N_{i}} \frac{1}{495 \times 436 \times 48} \sum_{h=1}^{495} \sum_{w=1}^{436} \sum_{k=1}^{48} (\hat{Y}(i, j, h, w, k) - Y(i, j, h, w, k))^{2},
    \label{equation_1}
\end{equation}

where $\hat{Y}(i, j, h, w, k)$ and $Y(i, j, h, w, k)$ denote the predicted and the ground truth pixel values in the $k$-th channel at position $(h, w)$ in the $j$-th training instance of the $i$-th city, respectively. The training instances are obtained using a sliding window over the daily traffic map movies. In the competition, the latest possible start time of the test slot is set to 8 p.m. We follow this and set the latest possible start time of the sliding window also to 8 p.m. Since for each hour we have $12$ heatmaps each corresponding to a 5-minute time bin, we thus have $12 \times (12 + 8) + 1 = 241$ training instances per city per day.

Multi-task learning has been widely adopted in computer vision [43], natural language processing [44], and recommender systems [45], and it is generally applied by sharing the hidden layers between all tasks, while keeping several task-specific output layers [46]. We experimented with adding city-specific layers to our multi-task learning model, but we found that it results in inferior performance. We presume that this is due to the strong homogeneity among the tasks of forecasting the future traffic flows in different cities.

The motivation behind multi-task learning is that, it allows the model to explore and exploit the shared knowledge in traffic forecasting for different cities. Since we concatenate the $9$-channel static information to the input, the multi-task learning framework would also encourage the model to capture the various graph structures within the road networks of different cities. We also observe that, due to the scarcity of the data, if we train on the data from one city only, the model quickly ``memorizes'' the training data and starts overfitting. Multi-task learning can be regarded as an implicit data augmentation and regularization technique [46], which explains its improved performance over the baseline approach. Furthermore, the multi-task learning framework enforces the model to learn city-agnostic representations, which significantly improves data efficiency and reduces overfitting.

\subsection{Implementation details}

All models are trained using the Adam optimizer [47] with $\beta_{1} = 0.9$, $\beta_{2} = 0.999$, batch size of $8$ and learning rate of $1e-4$. For the core challenge (temporal domain adaptation), we employ the U-Net architecture with $4$ downsampling layers and $4$ upsampling layers, and train the model for $5$ epochs. For the extended challenge (spatio-temporal domain adaptation), we employ the U-Net architecture with only $1$ downsampling layer and $1$ upsampling layer, and train the model for $50,000$ steps. Using less parameters and the early stopping strategy help the model prevent overfitting, which results in improved generalization capability when making predictions for entirely new cities.

\begin{table}[!t]
\centering
\begin{tabular}{C{0.75\textwidth}|C{0.15\textwidth}}
Training data & MSE \\ \hline \hline
\{Antwerp, Barcelona, Moscow\} \{2019, 2020\} data & 35.374 \\ \hline
Bangkok 2019 data & 35.118 \\ \hline
Bangkok 2019 data + \{Antwerp, Barcelona, Moscow\} 2019 data & 34.977 \\ \hline
Bangkok 2019 data + \{Antwerp, Barcelona, Moscow\} 2020 data & 34.826 \\ \hline
Bangkok 2019 data + Barcelona \{2019, 2020\} data & 34.609 \\ \hline
\textbf{Bangkok 2019 + \{Antwerp, Barcelona, Moscow\} \{2019, 2020\} data} & \textbf{34.315} \\ \hline
\end{tabular}
\caption{Further experiments on temporal domain adaptation where we use the in-Covid 2020 data of Bangkok as our local validation set and vary the setup of the training data.}
\label{table:core_3}
\end{table}

\section{Experiments}

\subsection{Dataset and evaluation metric}

The competition dataset consists of traffic map movies covering $10$ diverse cities around the world from $2019$ through $2020$, and there are three types of different cities:

\begin{enumerate}[(i)]
    \item For training, there are $4$ cities (Antwerp, Bangkok, Barcelona, and Moscow). For each city, we have a training set of $181$ full training days both for pre-Covid $2019$ and in-Covid $2020$, and we have no test set.
    \item For the core competition (temporal domain adaptation), there are $4$ cities (Berlin, Chicago, Istanbul, and Melbourne). For each city, we have a training set of $181$ full training days for pre-Covid $2019$ only, and the test set consists of traffic map movies for in-Covid $2020$.
    \item For the extended competition (spatio-temporal domain adaptation), there are $2$ cities (New York and Vienna). We have no training data for cities in the extended competition, and test set consists of traffic map movies both for pre-Covid $2019$ and in-Covid $2020$.
\end{enumerate}

The evaluation metric used in the challenge is the standard mean squared error (MSE) criterion, which is computed by averaging the pixel-wise squared difference between the predicted traffic map movies and the ground truths, where all cities and all channels are given equal weights; see Equation \ref{equation_1}.

\subsection{Baseline models}

\begin{table}[!t]
\centering
\begin{tabular}{L{0.38\textwidth}|C{0.2\textwidth}|C{0.22\textwidth}}
Method & MSE & Training time (hours) \\ \hline \hline
Naive Average & 63.140 & - \\ \hline
U-Net [41] & 60.114 & 2.6 \\ \hline
Graph ResNet [27] & 60.537 & 10.8 \\ \hline
U-Net + AdaBN [48] & 60.253 & 2.6 \\ \hline
U-Net + DaNN [32] & 60.077 & 2.8 \\ \hline
U-Net + DDC [33] & 60.274 & 3.0 \\ \hline
U-Net + DeepCORAL [34] & 60.361 & 2.8 \\ \hline
U-Net + ADDA [35] & 60.157 & 2.8 \\ \hline
U-Net + DANN [36] & 60.175 & 2.8 \\ \hline
U-Net + DSN [37] & 60.012 & 3.2 \\ \hline
U-Net + MAML [50] & 60.249 & 18.2 \\ \hline
\textbf{U-Net + Multi-task Learning (ours)} & \textbf{59.732} & \textbf{1.2} \\ \hline
\end{tabular}
\caption{Performance of our multi-task learning approach and the baseline methods in the extended competition.}
\label{table:extended_1}
\end{table}

We compare our multi-task learning approach against the following baseline methods:

\begin{enumerate}
  \item \textbf{Naive Average}: The naive average baseline takes the average of the input frame as the prediction for all lead times. 
  \item \textbf{U-Net} [41]: The baseline U-Net is trained and evaluated on each city separately. In the core competition (temporal domain adaptation), we train on the provided pre-Covid $2019$ data for the target city. In the extended competition (spatio-temporal domain adaptation), since no training data is provided for the target city, we train on the pre-Covid $2019$ and in-Covid $2020$ data for Bangkok.
  \item \textbf{Graph ResNet} [27]: Graph ResNet is a ResNet-inspired graph convolutional neural network (GCN) approach that have excellent generalization capability in traffic forecasting.
  \item \textbf{U-Net + AdaBN} [48]: Adaptive Batch Normalization (AdaBN) is a simple yet effective approach to domain adaptation which transfers the trained model to a new domain by modulating the statistics in the Batch Normalization layers.
  \item \textbf{U-Net + DaNN} [32]: Domain Adaptive Neural Networks (DaNN) reduces the distribution mismatch between the source and target domains in the latent space by adopting the Maximum Mean Discrepancy (MMD) measure as a regularization term.
  \item \textbf{U-Net + DDC} [33]: Deep Domain Confusion (DDC) introduces an adaptation layer and a domain confusion loss which enforce the model to learn semantically meaningful and domain invariant representations.
  \item \textbf{U-Net + Deep CORAL} [34]: Deep CORAL extends the CORrelation ALignment (CORAL) method [49] by learning a non-linear transformation with deep neural networks.
  \item \textbf{U-Net + ADDA} [35]: Adversarial Discriminative Domain Adaptation (ADDA) is an unsupervised adversarial domain adaptation approach which works by jointly learning discriminative representations using the labels in the source domain and learning a separate encoding that maps the data from the target domain to the same feature space through a domain-adversarial loss. 
  \item \textbf{U-Net + DANN} [36]: Domain-adversarial neural network (DANN) adapts a gradient reversal layer that enables adversarial domain adaptation through back-propagation.
  \item \textbf{U-Net + DSN} [37]: Domain Separation Networks (DSN) learns domain-invariant representations by explicitly separating the feature representations private to each domain from those that are shared between the domains.
  \item \textbf{U-Net + MAML} [50]: Model-Agnostic Meta-Learning (MAML) is a state-of-the-art meta-learning algorithm that learns the weights of deep neural networks through gradient descent. 
\end{enumerate}

We follow previous literature to set up the hyperparameters and training strategies for the baseline models. For DaNN, DDC, Deep CORAL, ADDA, DANN, and DSN, we empirically apply the domain adaptation losses to the output of the penultimate block of the decoder in the U-Net since it results in best performance in our experiments. We ran our multi-task learning approach and all baseline methods $3$ times with different random seeds and report the average leaderboard score for each model.

\subsection{Temporal domain adaptation}

The average leaderboard score for each model in the core competition (temporal domain adaptation) is presented in Table \ref{table:core_1}. We observe that several domain adaptation approaches, e.g., DaNN, DSN, and MAML, are able to achieve better temporal domain adaptation performance than the baseline U-Net, which is trained and evaluated on each city separately. Meanwhile, our multi-task learning approach achieves the best performance, outperforming all the baseline domain adaptation methods. In addition, the mutli-task learning approach is highly efficient in terms of the training time since it only requires training one model for all cities.

We present the final leaderboard result for the core competition \footnote{https://www.iarai.ac.at/traffic4cast/competitions/t4c-2021-core-temporal/?leaderboard} in Table \ref{table:core_2}. Our final leaderboard result is obtained by an ensemble of U-Net models trained with multi-task learning, where we average the predictions from models trained with different random seeds, and it achieves the $1$st place among all the participating teams. We tried more sophisticated ensembling techniques such as weighted averaging [51] or amplifying the relative difference through exponents [52], but did not see gains as expected.

In order to gain a better understanding of the effectiveness of our multi-task learning approach, we conduct further experiments where we use the in-Covid 2020 data of Bangkok as our local validation set and vary the setup of training data; see Table \ref{table:core_3}. We observe that obtaining training data from the same city (Bangkok 2019 data) is more important than collecting more data from different cities. This is not surprising since different cities can have significantly different traffic patterns. In addition, comparing the training data from different cities, the 2020 data brings more performance gains than the 2019 data, since it has a more similar pattern with the validation set. Finally, adding the 2019 and 2020 data from one city (Barcelona) to the training set is more effective than adding the 2019 or 2020 data from three cities (Antwerp, Barcelona, and Moscow), despite the less amount of training data. This suggests that the multi-task learning approach not only works because of the implicit data augmentation, but also because when the training set consists of data from both 2019 and 2020, multi-task learning encourages the model to learn to adapt to temporal domain shifts during training. 

\subsection{Spatio-temporal domain adaptation}

\begin{table}[!t]
\centering
\begin{tabular}{C{0.2\textwidth}|C{0.2\textwidth}|C{0.2\textwidth}}
Rank & Team & MSE \\ \hline \hline
1 & sungbin & 59.559 \\ \hline
\textbf{2} & \textbf{oahciy (ours)} & \textbf{59.586} \\ \hline
3 & nina & 59.915 \\ \hline
4 & dninja & 60.221\\ \hline
5 & HBKU & 60.266 \\ \hline
\end{tabular}
\caption{Extended competition leaderboard results for the top $5$ teams.}
\label{table:extended_2}
\end{table}

The average leaderboard score for each model in the extended competition (spatio-temporal domain adaptation) is presented in Table \ref{table:extended_1}. We observe that our baseline U-Net is able to outperform the Graph ResNet approach, which was reported to achieve state-of-the-art performance in spatio-temporal domain adaptation [27]. This suggests that reducing the number of parameters in the U-Net model, together with adopting an early stopping strategy, can effectively prevent the model from overfitting, which leads to better spatio-temporal domain adaptation performance. In addition, several domain adaptation approaches are able to outperform the baseline U-Net model trained and evaluated on each city separately, albeit by a small margin. Our multi-task learning approach again achieves the best performance among all the compared methods. The multi-task learning approach is also highly efficient, since it only requires training one model for all cities and does not introduce extra computation costs.

We present the final leaderboard result for the extended competition \footnote{https://www.iarai.ac.at/traffic4cast/competitions/t4c-2021-extended-spatiotemporal/?leaderboard} in Table \ref{table:extended_2}. Our final leaderboard result is also obtained by an ensemble of U-Net models trained with multi-task learning, and it achieves the $2$nd place among all the participating teams, with less than $0.05\%$ relative difference from the $1$st place team.

\section{Discussion and future work}

Since our purpose was to find an effective approach to temporal and spatio-temporal domain adaptation for traffic forecasting models, we did not spend much time in exploring the optimal U-Net architecture in this challenge. Therefore we can potentially achieve even better performance by utilizing more sophisticated U-Net architectures [53]. Alternatively, we can employ Transformer-based vision models [54, 55, 56], which have shown superior performance in a number of computer vision tasks, or other state-of-the-art spatio-temporal learning models [3, 21, 22]. 

In addition, our results were achieved by using the same trained network weights for all cities. Apparently, the optimal performance for each city would not be achieved at the same time, and better performance can be achieved if we test the performance on each city in an trial-and-error fashion. We deliberately chose not to fit different models for different cities so as not to exploit the leaderboard. Nevertheless, in real-world scenarios, it is worth experimenting with a per-city optimization of the multi-task learning approach.

Last but not least, our solution to this competition is based on a purely data-driven approach, and we did not perform feature engineering. Feature engineering with domain expertise has been the key to winning in various machine learning competitions [57, 58, 59, 60]. We can expect a further performance boost if we incorporate a few manually designed features, e.g., the location of the pixel in the heatmap, the time of day for the predicted traffic flows, and whether it is a weekend or a holiday.

\section{Conclusion}

We present a multi-task learning framework for traffic forecasting. Experimental results demonstrate that our multi-task learning approach achieves strong empirical performance, outperforming a number of baseline domain adaptation methods, while remaining highly efficient. 

\section*{References}
{
\small
[1] Boukerche, Azzedine, and Jiahao Wang. "Machine Learning-based traffic prediction models for Intelligent Transportation Systems." Computer Networks 181 (2020): 107530.

[2] Lv, Yisheng, Yanjie Duan, Wenwen Kang, Zhengxi Li, and Fei-Yue Wang. "Traffic flow prediction with big data: a deep learning approach." IEEE Transactions on Intelligent Transportation Systems 16, no. 2 (2014): 865-873.

[3] Yu, Wei, Yichao Lu, Steve Easterbrook, and Sanja Fidler. "Efficient and Information-Preserving Future Frame Prediction and Beyond." In International Conference on Learning Representations. 2019.

[4] Shelhamer, Evan, Kate Rakelly, Judy Hoffman, and Trevor Darrell. "Clockwork convnets for video semantic segmentation." In European Conference on Computer Vision, pp. 852-868. Springer, Cham, 2016.

[5] Shi, Xingjian, Zhihan Gao, Leonard Lausen, Hao Wang, Dit-Yan Yeung, Wai-kin Wong, and Wang-chun WOO. "Deep Learning for Precipitation Nowcasting: A Benchmark and A New Model." Advances in Neural Information Processing Systems 30 (2017): 5617-5627.

[6] Kreil, David P., Michael K. Kopp, David Jonietz, Moritz Neun, Aleksandra Gruca, Pedro Herruzo, Henry Martin, Ali Soleymani, and Sepp Hochreiter. "The surprising efficiency of framing geo-spatial time series forecasting as a video prediction task–Insights from the IARAI Traffic4cast Competition at NeurIPS 2019." In NeurIPS 2019 Competition and Demonstration Track, pp. 232-241. PMLR, 2020.

[7] Kopp, Michael, David Kreil, Moritz Neun, David Jonietz, Henry Martin, Pedro Herruzo, Aleksandra Gruca et al. "Traffic4cast at NeurIPS 2020? yet more on theunreasonable effectiveness of gridded geo-spatial processes." In NeurIPS 2020 Competition and Demonstration Track, pp. 325-343. PMLR, 2021.

[8] Dimitrakopoulos, George, and Panagiotis Demestichas. "Intelligent transportation systems." IEEE Vehicular Technology Magazine 5, no. 1 (2010): 77-84.

[9] Dong, Honghui, Limin Jia, Xiaoliang Sun, Chenxi Li, and Yong Qin. "Road traffic flow prediction with a time-oriented ARIMA model." In 2009 Fifth International Joint Conference on INC, IMS and IDC, pp. 1649-1652. IEEE, 2009.s

[10] Huang, Da-Rong, Jun Song, Da-Cheng Wang, Jian-Qiu Cao, and Wei Li. "Forecasting model of traffic flow based on ARMA and wavelet transform." Computer Engineering and Applications 42, no. 36 (2006): 191-194.

[11] Yang, Wen, Dongyuan Yang, Yali Zhao, and Jinli Gong. "Traffic flow prediction based on wavelet transform and radial basis function network." In 2010 International Conference on Logistics Systems and Intelligent Management (ICLSIM), vol. 2, pp. 969-972. IEEE, 2010.

[12] Yi, Hongsuk, HeeJin Jung, and Sanghoon Bae. "Deep neural networks for traffic flow prediction." In 2017 IEEE international conference on big data and smart computing (BigComp), pp. 328-331. IEEE, 2017.

[13] Chen, Cen, Kenli Li, Sin G. Teo, Xiaofeng Zou, Keqin Li, and Zeng Zeng. "Citywide traffic flow prediction based on multiple gated spatio-temporal convolutional neural networks." ACM Transactions on Knowledge Discovery from Data (TKDD) 14, no. 4 (2020): 1-23.

[14] Yu, Wei, Yichao Lu, Steve Easterbrook, and Sanja Fidler. "CrevNet: Conditionally Reversible Video Prediction." arXiv preprint arXiv:1910.11577 (2019).

[15] Sun, Shangyu, Huayi Wu, and Longgang Xiang. "City-wide traffic flow forecasting using a deep convolutional neural network." Sensors 20, no. 2 (2020): 421.

[16] Guo, Shengnan, Youfang Lin, Shijie Li, Zhaoming Chen, and Huaiyu Wan. "Deep spatial–temporal 3D convolutional neural networks for traffic data forecasting." IEEE Transactions on Intelligent Transportation Systems 20, no. 10 (2019): 3913-3926.

[17] Hochreiter, Sepp, and Jürgen Schmidhuber. "Long short-term memory." Neural computation 9, no. 8 (1997): 1735-1780.

[18] Cho, Kyunghyun, Bart van Merriënboer, Caglar Gulcehre, Dzmitry Bahdanau, Fethi Bougares, Holger Schwenk, and Yoshua Bengio. "Learning Phrase Representations using RNN Encoder–Decoder for Statistical Machine Translation." In Proceedings of the 2014 Conference on Empirical Methods in Natural Language Processing (EMNLP), pp. 1724-1734. 2014.

[19] Fu, Rui, Zuo Zhang, and Li Li. "Using LSTM and GRU neural network methods for traffic flow prediction." In 2016 31st Youth Academic Annual Conference of Chinese Association of Automation (YAC), pp. 324-328. IEEE, 2016.

[20] Xingjian, S. H. I., Zhourong Chen, Hao Wang, Dit-Yan Yeung, Wai-Kin Wong, and Wang-chun Woo. "Convolutional LSTM network: A machine learning approach for precipitation nowcasting." In Advances in neural information processing systems, pp. 802-810. 2015.

[21] Wang, Yunbo, Mingsheng Long, Jianmin Wang, Zhifeng Gao, and Philip S. Yu. "Predrnn: Recurrent neural networks for predictive learning using spatiotemporal lstms." In Proceedings of the 31st International Conference on Neural Information Processing Systems, pp. 879-888. 2017.

[22] Wang, Yunbo, Zhifeng Gao, Mingsheng Long, Jianmin Wang, and S. Yu Philip. "Predrnn++: Towards a resolution of the deep-in-time dilemma in spatiotemporal predictive learning." In International Conference on Machine Learning, pp. 5123-5132. PMLR, 2018.

[23] Bui, Khac-Hoai Nam, Jiho Cho, and Hongsuk Yi. "Spatial-temporal graph neural network for traffic forecasting: An overview and open research issues." Applied Intelligence (2021): 1-12.

[24] Zhao, Ling, Yujiao Song, Chao Zhang, Yu Liu, Pu Wang, Tao Lin, Min Deng, and Haifeng Li. "T-gcn: A temporal graph convolutional network for traffic prediction." IEEE Transactions on Intelligent Transportation Systems 21, no. 9 (2019): 3848-3858.

[25] Zheng, Chuanpan, Xiaoliang Fan, Cheng Wang, and Jianzhong Qi. "Gman: A graph multi-attention network for traffic prediction." In Proceedings of the AAAI Conference on Artificial Intelligence, vol. 34, no. 01, pp. 1234-1241. 2020.

[26] Zhou, Fan, Qing Yang, Ting Zhong, Dajiang Chen, and Ning Zhang. "Variational graph neural networks for road traffic prediction in intelligent transportation systems." IEEE Transactions on Industrial Informatics 17, no. 4 (2020): 2802-2812.

[27] Martin, Henry, Dominik Bucher, Ye Hong, René Buffat, Christian Rupprecht, and Martin Raubal. "Graph-ResNets for short-term traffic forecasts in almost unknown cities." In NeurIPS 2019 Competition and Demonstration Track, pp. 153-163. PMLR, 2020.

[28] Pan, Sinno Jialin, and Qiang Yang. "A survey on transfer learning." IEEE Transactions on knowledge and data engineering 22, no. 10 (2009): 1345-1359.

[29] Pan, Sinno Jialin, Ivor W. Tsang, James T. Kwok, and Qiang Yang. "Domain adaptation via transfer component analysis." IEEE transactions on neural networks 22, no. 2 (2010): 199-210.

[30] Zhang, Yuchen, Tianle Liu, Mingsheng Long, and Michael Jordan. "Bridging theory and algorithm for domain adaptation." In International Conference on Machine Learning, pp. 7404-7413. PMLR, 2019.

[31] Wang, Mei, and Weihong Deng. "Deep visual domain adaptation: A survey." Neurocomputing 312 (2018): 135-153.

[32] Ghifary, Muhammad, W. Bastiaan Kleijn, and Mengjie Zhang. "Domain adaptive neural networks for object recognition." In Pacific Rim international conference on artificial intelligence, pp. 898-904. Springer, Cham, 2014.

[33] Tzeng, Eric, Judy Hoffman, Ning Zhang, Kate Saenko, and Trevor Darrell. "Deep domain confusion: Maximizing for domain invariance." arXiv preprint arXiv:1412.3474 (2014).

[34] Sun, Baochen, and Kate Saenko. "Deep coral: Correlation alignment for deep domain adaptation." In European conference on computer vision, pp. 443-450. Springer, Cham, 2016.

[35] Tzeng, Eric, Judy Hoffman, Kate Saenko, and Trevor Darrell. "Adversarial discriminative domain adaptation." In Proceedings of the IEEE conference on computer vision and pattern recognition, pp. 7167-7176. 2017.

[36] Ganin, Yaroslav, Evgeniya Ustinova, Hana Ajakan, Pascal Germain, Hugo Larochelle, François Laviolette, Mario Marchand, and Victor Lempitsky. "Domain-adversarial training of neural networks." The journal of machine learning research 17, no. 1 (2016): 2096-2030.

[37] Bousmalis, Konstantinos, George Trigeorgis, Nathan Silberman, Dilip Krishnan, and Dumitru Erhan. "Domain separation networks." Advances in neural information processing systems 29 (2016): 343-351.

[38] Zhang, Jing, Wanqing Li, and Philip Ogunbona. "Unsupervised domain adaptation: A multi-task learning-based method." Knowledge-Based Systems 186 (2019): 104975.

[39] Yang, Yongxin, and Timothy M. Hospedales. "A unified perspective on multi-domain and multi-task learning." arXiv preprint arXiv:1412.7489 (2014).

[40] Zhou, Fan, Brahim Chaib-draa, and Boyu Wang. "Multi-task Learning by Leveraging the Semantic Information." In Proceedings of the AAAI Conference on Artificial Intelligence, vol. 35, no. 12, pp. 11088-11096. 2021.

[41] Ronneberger, Olaf, Philipp Fischer, and Thomas Brox. "U-net: Convolutional networks for biomedical image segmentation." In International Conference on Medical image computing and computer-assisted intervention, pp. 234-241. Springer, Cham, 2015.

[42] Wu, Yuxin, and Kaiming He. "Group normalization." In Proceedings of the European conference on computer vision (ECCV), pp. 3-19. 2018.

[43] Li, Xi, Liming Zhao, Lina Wei, Ming-Hsuan Yang, Fei Wu, Yueting Zhuang, Haibin Ling, and Jingdong Wang. "Deepsaliency: Multi-task deep neural network model for salient object detection." IEEE transactions on image processing 25, no. 8 (2016): 3919-3930.

[44] Worsham, Joseph, and Jugal Kalita. "Multi-task learning for natural language processing in the 2020s: where are we going?." Pattern Recognition Letters 136 (2020): 120-126.

[45] Lu, Yichao, Ruihai Dong, and Barry Smyth. "Why I like it: multi-task learning for recommendation and explanation." In Proceedings of the 12th ACM Conference on Recommender Systems, pp. 4-12. 2018.

[46] Ruder, Sebastian. "An overview of multi-task learning in deep neural networks." arXiv preprint arXiv:1706.05098 (2017).

[47] Kingma, Diederik P., and Jimmy Ba. "Adam: A Method for Stochastic Optimization." In ICLR (Poster). 2015.

[48] Li, Yanghao, Naiyan Wang, Jianping Shi, Jiaying Liu, and Xiaodi Hou. "Revisiting batch normalization for practical domain adaptation." arXiv preprint arXiv:1603.04779 (2016).

[49] Sun, Baochen, Jiashi Feng, and Kate Saenko. "Correlation alignment for unsupervised domain adaptation." In Domain Adaptation in Computer Vision Applications, pp. 153-171. Springer, Cham, 2017.

[50] Finn, Chelsea, Pieter Abbeel, and Sergey Levine. "Model-agnostic meta-learning for fast adaptation of deep networks." In International Conference on Machine Learning, pp. 1126-1135. PMLR, 2017.

[51] Lu, Yichao, Cheng Chang, Himanshu Rai, Guangwei Yu, and Maksims Volkovs. "Multi-view scene graph generation in videos." In International Challenge on Activity Recognition (ActivityNet) CVPR 2021 Workshop, vol. 3. 2021.

[52] Lu, Yichao, Ruihai Dong, and Barry Smyth. "Context-aware sentiment detection from ratings." In International Conference on Innovative Techniques and Applications of Artificial Intelligence, pp. 87-101. Springer, Cham, 2016.

[53] Zhou, Zongwei, Md Mahfuzur Rahman Siddiquee, Nima Tajbakhsh, and Jianming Liang. "Unet++: A nested u-net architecture for medical image segmentation." In Deep learning in medical image analysis and multimodal learning for clinical decision support, pp. 3-11. Springer, Cham, 2018.

[54] Dosovitskiy, Alexey, Lucas Beyer, Alexander Kolesnikov, Dirk Weissenborn, Xiaohua Zhai, Thomas Unterthiner, Mostafa Dehghani et al. "An Image is Worth 16x16 Words: Transformers for Image Recognition at Scale." In International Conference on Learning Representations. 2020.

[55] Liu, Ze, Yutong Lin, Yue Cao, Han Hu, Yixuan Wei, Zheng Zhang, Stephen Lin, and Baining Gao. "Swin Transformer: Hierarchical Vision Transformer Using Shifted Windows." In Proceedings of the IEEE/CVF International Conference on Computer Vision, pp. 10012-10022. 2021.

[56] Lu, Yichao, Himanshu Rai, Jason Chang, Boris Knyazev, Guangwei Yu, Shashank Shekhar, Graham W. Taylor, and Maksims Volkovs. "Context-aware Scene Graph Generation with Seq2Seq Transformers." In Proceedings of the IEEE/CVF International Conference on Computer Vision, pp. 15931-15941. 2021.

[57] Volkovs, Maksims, Himanshu Rai, Zhaoyue Cheng, Ga Wu, Yichao Lu, and Scott Sanner. "Two-stage model for automatic playlist continuation at scale." In Proceedings of the ACM Recommender Systems Challenge 2018, pp. 1-6. 2018.

[58] Volkovs, Maksims, Anson Wong, Zhaoyue Cheng, Felipe Pérez, Ilya Stanevich, and Yichao Lu. "Robust contextual models for in-session personalization." In Proceedings of the Workshop on ACM Recommender Systems Challenge, pp. 1-5. 2019.

[59] Lu, Yichao, Cheng Chang, Himanshu Rai, Guangwei Yu, and Maksims Volkovs. "Learning Effective Visual Relationship Detector on 1 GPU." arXiv preprint arXiv:1912.06185 (2019).

[60] Jannach, Dietmar, Gabriel de Souza P. Moreira, and Even Oldridge. "Why are deep learning models not consistently winning recommender systems competitions yet? A position paper." In Proceedings of the Recommender Systems Challenge 2020, pp. 44-49. 2020.
}
\end{document}